\pdfoutput=1

\documentclass[11pt]{article}

\usepackage{acl}

\usepackage{graphicx}
\usepackage{tabularx}
\usepackage{multirow}
\usepackage{siunitx}  
\usepackage{times}
\usepackage{latexsym}
\usepackage{amsmath}

\usepackage[T1]{fontenc}

\usepackage[utf8]{inputenc}
\usepackage{enumitem}
\setlist[itemize]{noitemsep, nolistsep}
\usepackage{balance}

\newcommand{\shrink}{\vspace*{-.9\baselineskip}}

\usepackage{microtype}

\usepackage{inconsolata}

\title{Generate then Refine: Data Augmentation for Zero-shot Intent Detection}

\author{I-Fan Lin \\
  Leiden University \\
  Leiden,  The Netherlands \\
  \texttt{\fontsize{9pt}{10pt}\selectfont i.lin@liacs.leidenuniv.nl} \\\And
  Faegheh Hasibi \\
  Radboud University \\
  Nijmegen, The Netherlands\\
  \texttt{\fontsize{9pt}{10pt}\selectfont faegheh.hasibi@ru.nl} \\\\\And
  Suzan Verberne \\
  Leiden University \\
  Leiden,  The Netherlands\\
\texttt{\fontsize{9pt}{10pt}\selectfont s.verberne@liacs.leidenuniv.nl} \\}

\begin{document}
\maketitle
\begin{abstract}
In this short paper we propose a data augmentation method for intent detection in zero-resource domains.
Existing data augmentation methods rely on few labelled examples for each intent category, which can be expensive in settings with many possible intents. 
We use a two-stage approach: First, we generate utterances for intent labels using an open-source large language model in a zero-shot setting. Second, we develop a smaller sequence-to-sequence model (the Refiner), to improve the generated utterances. The Refiner is fine-tuned on seen domains and then applied to unseen domains. We evaluate our method by training an intent classifier on the generated data, and evaluating it on real (human) data.
We find that the Refiner significantly improves the data utility and diversity over the zero-shot LLM baseline for unseen domains and over common baseline approaches.
Our results indicate that a two-step approach of a generative LLM in zero-shot setting and a smaller sequence-to-sequence model can provide high-quality data for intent detection.\footnote{Our code is available at: \url{https://github.com/tom192180/Generate_then_Refine} 
}
\end{abstract}

\section{Introduction}

Intent detection is 
a common component in task-oriented dialogue (TOD) systems. Its objective is to categorize user utterances into predefined classes of user intents \cite{ni2023recent}. The advent of transformer-based models has significantly elevated the performance of intent detection, particularly when trained and evaluated within familiar domains and intents \cite{larson2019evaluation}. However, when faced with previously unseen domains and intents, a considerable challenge emerges, primarily stemming from data scarcity.

Addressing this challenge involves maximizing the utility of limited training data and ensuring the adaptability of the trained intent classifier to novel intents. A combination of few-shot learning and meta-learning strategies has been employed in prior work 
\cite{zhang2020discriminative,sauer-etal-2022-knowledge,wang2023dual, cho2023prompt}. 

\begin{figure}[t]
  \centering
  \includegraphics[width=\linewidth]{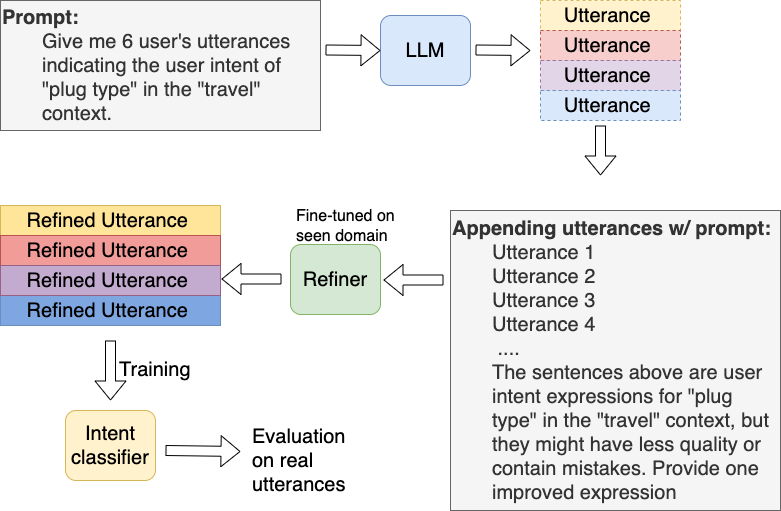}
  \shrink
  \caption{The pipeline of generating and refining utterances for out-of-domain intent detection.}
  \label{fig:architecture}
  \shrink
\end{figure}

A viable alternative solution for low-resource domain adaptation is augmenting and increasing the training data. 
This is achieved by  direct utilization of a generative large language model (LLM) as a data generator \cite{xia2020cg, lee2021neural, sahu2022data, marceau2022quick, fang2023chatgpt}. 
Synthetically generated data, however, is not always of high quality and this has direct effect on model performance. As a remedy,  
several works have attempted to filter out generated examples that are of low quality or relevance 
\cite{sahu2022data, meng2022generating, lin2023selective}.

In this paper, we study the setup where no labelled data is available for unseen domains (e.g., travel), while substantial number of user utterances with intent labels exist for seen domains (e.g., banking).
Our goal is to create high-quality training data to train an intent classifier for completely unseen domains. In these unseen domains, the intent labels are known, but there are no utterances available. 
In line with previous work \cite{ye2022zerogen, sahu2022data, meng2022generating, lin2023selective}, we obtain generated data from a generative LLM. The utility of this data, as measured by classification quality,  lags behind a classifier trained on human utterances.
One of the causes is that the LLM tends to generate the words from the intent label in the output utterances, which makes it less representative of real user's utterances. 
Therefore, we propose a Refiner model that transforms sub-optimal generated utterances into better ones. The goal of our Refiner model, trained on data from seen domains, is  to enhance the performance of the intent classifier; see Figure~\ref{fig:architecture} for illustration. 




The main contribution of this paper is proposing a sequence-to-sequence learning method for enhancing the utility of LLM-generated utterances, while retaining sample size. This stands in contrast to the previous studies that utilize formula-based metrics for data selection to filter sub-optimal utterances \cite{meng2022generating, lin2023selective}. 

In summary, the contributions of this paper are:
\begin{itemize}
    \item Proposing a Refiner model for zero-shot intent classification and showing its effectiveness compared to state-of-the-art data augmentation approaches, including ChatGPT. 
    \item Evaluating the Refiner (via extensive analysis and ablation study) as a compute-efficient model, showing that a smaller model, when trained on rich-resource domains, is capable of improving the output of a larger (7B-parameter) LLM in unseen domains.
    \item Showing the success of the Refiner in producing lexically diverse text, comparable to human data, thereby addressing the common issue of less diverse text generated by LLMs.
\end{itemize}

\section{Related Work}



Data augmentation is a commonly used solution to data scarcity in training conversational agents \cite{soudani2024survey}. 
Recently, 
generating new examples directly from LLMs has emerged as a viable strategy \cite{meng2022generating, fang2023chatgpt, dai2023chataug, yu2024large, guo2024generative}.
In addressing the challenge of data scarcity for intent detection in novel domains, some of the prior work has  applied metric-based meta-learning algorithms \cite{zhang2022mgimn, sauer-etal-2022-knowledge, wang2023dual}. Data augmentation is an alternative. \citet{lee2021neural} employ extrapolation techniques with a sequence-to-sequence model to generate new, under-represented utterances. 
Alternatively, leveraging the advancements in LLMs, \citet{sahu2022data} use GPT-3 with in-context learning to get additional utterances. \citet{lin2023selective} adopted a similar appraoch, using LLMs 
to generate synthetic examples, and then selecting  useful data for training based on the Pointwise V-Information metric. 
Our approach differs from previous research in two key ways. Firstly, 
we specifically consider the zero-shot (instead of the few-shot) setting. This increases the challenge for LLMs to provide high-quality data. Secondly, rather than relying on a formula-based metric, we opt for a potentially more flexible approach -- the Refiner. The Refiner directly generates refined data instead of filtering out irrelevant samples. 



\section{Method}
\subsection{Task Formulation}
We assume the labeled dataset $D = \{(u_{i}, y_{i})|y_{i} \in Y_s\}_{i=1}^{N}$, where $u_i$ denotes $i_{th}$ utterance labelled with intent $y_i$, $Y_s$ denotes a set of distinct intents for seen domains, and $N$ is the number of labelled utterances. Assuming that $D$ provides sufficient amount of training data for seen domains (i.e., $N$ is large), our objective is to predict the unseen intent class set $Y_{t}$, where $Y_{s} \cap Y_{t} = \emptyset $ and $t$ denotes unseen domains. 
We consider the zero-shot setting, assuming that we have no example utterances in the unseen domain. 
\subsection{Utterance Generation with LLMs}
We use generative LLMs to generate utterances for unseen intents. 
We prompt the model directly with a given intent in a zero-shot setting, requesting it to provide corresponding utterances (see Figure~\ref{fig:architecture}). The zero-shot use of an LLM for this task leads to suboptimal utterances that cannot directly be used as a replacement for human-generated data to train the intent classifier \cite{ye2022zerogen}. 
To improve the quality of the generated utterances, we use training data from seen domains with sufficient labelled utterances. We train an utterance Refiner: a sequence-to-sequence model that takes generated utterances as input and outputs their refined versions. 
After training the Refiner on seen domains, it is applied to LLM-generated utterances from unseen domains to improve them.

\subsection{Refiner}

\textbf{Refiner goal.} The objective of the Refiner is to generate diverse, high-quality utterances from lower-quality inputs. We anticipate that the Refiner can effectively fulfill two roles:
    (a) In the case of irrelevant or inaccurate utterances generated by the LLM, the Refiner should be able to generate utterances accurately aligned with the intended intents.
    (b) The generation of a diverse set of utterances holds the promise of significantly improving the overall performance of the intent classifier.





\textbf{Training objective.} Let $\mathcal{U}_{j}^{gen}$ denote the set of generated utterances and $\mathcal{U}_{j}^{real}$ denote the set of human-written utterances for the $j_{th}$ unique intent from the seen domain. $\mathcal{U}_{j}^{gen}$ is generated using an LLM, while $\mathcal{U}_{j}^{real}$ is derived from the provided labeled dataset. We ensure that the length of the generated set matches that of our labeled dataset, i.e., $|\mathcal{U}_{j}^{gen}| =|\mathcal{U}_{j}^{real}| = N_j$. For each $j_{th}$ distinct intent from the seen domain, a sub-training set $\hat{D_{j}}$ is defined as
 $\hat{D_{j}} = \{ (U^{gen}_{ij}, U^{real}_{ij})\}_{i=1}^{N_j}$, 
where $U^{gen}_{ij} \subset \mathcal{U}^{gen}_{j} $ and $U^{real}_{ij} \subset \mathcal{U}^{real}_{j}$; and $\sum_{j=1}^{k} N_j = N$, where $k$ is the total number of distinct intents. The complete training set for the Refiner is then represented as
$\hat{D} = \bigcup_{j=1}^{k} \hat{D_{j}}$. 
%


To ensure our Refiner generalizes well, we select some domains from seen domains as validation set, using the remaining seen domains for training. We use the regular loss function of the sequence-to-sequence model for training. This loss evaluates to what extent the refined utterance (the model's output) is different from the real utterance used as ground truth. As validation loss, we use a distinct metric based on classification loss. The classifier is fine-tuned with the labeled dataset ${D}$ and is used to monitor the Refiner's performance during training with the validation domains.


\section{Experimental Settings}

\textbf{Datasets.} We use the CLINC150 \cite{larson2019evaluation} and SGD \cite{rastogi2020towards} datasets for our experiments. \textbf{CLINC150} includes 10 domains, with a total of 150 intents (15 per domain). 
Each intent has 100 training examples and 30 test examples. \textbf{SGD} includes 20 domains, with a total of 46 intents. Intents in SGD have a diverse number of utterances, ranging from hundreds to thousands. The domains included in both datasets are detailed in Appendix ~\ref{sec:domain}.

\textbf{LLMs for utterances generation.} We experiment with two LLMs to generate utterances: \textbf{Zephyr-7B Beta} \cite{tunstall2023zephyr} and \textbf{Llama3-8B-Instruct} \cite{llama3modelcard}.
An example of the prompt we use for zero-shot utterance generation is shown in Figure~\ref{fig:architecture}
 (top left).
\noindent\hfill\break

\textbf{Training the Refiner.} Flan-T5-large \cite{chung2024scaling} with 783M parameters serves as the backbone model for the Refiner. Pairs of seen-domain utterances and LLM-generated utterances are used to train the Refiner, which is trained for 6 epochs with a batch size of 24 and early stopping to prevent overfitting. During training, validation loss is monitored using a fine-tuned DistilBERT \cite{Sanh2019DistilBERTAD}\footnote{\url{https://huggingface.co/distilbert/distilbert-base-uncased}} classifier (67M parameters), trained on data from all seen domains. The classifier is fine-tuned over 1,800 steps with a batch size of 60, also using early stopping to avoid overfitting.
The prompt we use for the Refiner is:

 \begin{quote}The sentences above are user intent expressions for ``\{\textit{intent}\}'' in the ``\{\textit{domain}\}'' context, but they might have less quality or contain mistakes. Provide one improved expression.\end{quote}

Here, ``\{\textit{intent}\}'' and ``\{\textit{domain}\}'' are placeholders derived from the datasets.

\textbf{Data sampling for Refiner training.} To allow the Refiner to observe the diversity of the sample as a whole instead of a single input, we experiment with a multiple-to-one setting, aligning with previous research \cite{lee2021neural}. Based on experimentation with different values (See the ablation study in the Section~\ref{sec:abstudy}), we set the input number to 7 during both training and inference. This includes the current utterance plus 6 randomly selected utterances, with sampling done with replacement.  The results of Refiner are obtained with the 7to1 setting (unless otherwise specified).

\textbf{Evaluation.} The main criterion for the generated utterances is \textit{utility}: to what extent can the same task be performed with the generated utterances as with real (human) utterences. We therefore report on the accuracy of intent classification evaluation to assess the efficacy of the refined synthetic utterances. For this, we fine-tune DistilBERT\footnotemark[2] on the generated utterances with intent labels. 
The evaluation is conducted on authentic test data (real user utterances). To make the experiment results more generalized, for every experiment, we randomly split the domains into ``seen'' and ``unseen'', and perform 5 experiments with different splits. The evaluation metrics are based on the average of these five runs. For training the intent classifier, for each unseen-domain intent, we use 100 generated utterances for \textbf{CLINC150} and 200 for \textbf{SGD}. All reported accuracy results share the same classifier setting: a batch size of 60 and 1,800 training steps. The  utterances were split into train and validation sets with an 80\%-20\% ratio. Early stopping was applied to prevent overfitting.

For intrinsic evaluation of the generated data, we report on diversity (distinct-1 \& 2) \citep{li2016diversity, nakamura2019another} and account for penalization of longer text in computing distinct-n, following \cite{Joko:2024:LAPS}; see Appendix~\ref{sec:eval} for details about implementing distinct-n.  

\textbf{Train-test split} For the \textbf{CLINC150} dataset, we randomly select 5 out of the 10 domains as unseen domains for each experiment. We train on 4 seen domains and monitor validation loss on 1 seen domain, using a total of 7,500 examples ($= 100 * 15 * 5$), and evaluate on the remaining 5 unseen domains (75 intents), with a total of 2,250 test examples ($= 30 * 15 * 5$). For the \textbf{SGD} dataset,  to generalize the results and align with our task formulation (details in Appendix ~\ref{sec:split}), we merge all splits and randomly select 8 out of 20 domains as unseen domains. We train on 9 seen domains and monitor validation loss on 3 seen domains. Table \ref{tab:SGD_Splitdata} provides detailed splits for each experiment.

\begin{table}[t]
  \centering
  \small
   \begin{tabular}{|c|c|c|c|c|c|}
    \hline
    Experiment ID   & \# seen intents & \# unseen intents\\
     \hline
    1 & 28 (5.3K) & 18 (22.7K) \\
    \hline
    2 &  26  (5.1K)  &20 (21,6K) \\
    \hline
    3 &  29   (5.7K) &17 (16,8K) \\

    \hline
    4 &  25   (4.8K) &21 (26,7K) \\
    \hline
    5  &  29   (5.6K) &17 (18,3K)\\
    \hline

  \end{tabular}
  \shrink
  \caption{SGD train and test split: the number in parentheses indicates the total training and testing examples.}
  \label{tab:SGD_Splitdata}
\end{table}

\textbf{Baselines.} We compare our method to \citet{ye2022zerogen}'s \textbf{ZeroGen}, utilizing data directly generated from the LLM for downstream tasks. We also compare  with \citet{meng2022generating}'s \textbf{SuperGen} data selection approach, which performs data augmentation in the zero-shot setting by extensively sampling from an LLM and selecting a subset based on output confidence (see for more details Appendix~\ref{sec:baseline}). In both baselines we use the same generative LLMs as in our own method (Zephyr and Llama-3). Because \textbf{GPT-3.5-turbo} \cite{openai2023chatgpt}, referred to as ChatGPT, is widely used and shows strong zero-shot performance for many tasks, we use it as an additional baseline in the ZeroGen setting.
\footnote{We did not try SuperGen's data selection approach for ChatGPT because it requires oversampling and would incur ten times the cost of the API.} For all baselines and our approach, we use the same prompt to generated synthetic utterances.


\textbf{Hardware/resources used.} One NVIDIA A100 GPU, equipped with 40GB of memory, was used in all experiments. Additionally, it cost approximately 4 USD for using the ChatGPT API.

\section{Results}

\begin{table}[t]
  \centering
  \small
   \begin{tabular}{|l|c|c|c|c|c|}
    \hline
     & SGD   & Clinc150\\
      \hline
    ZeroGen (Zephyr) &  61.3 (4.9) &  69.9 (2.0)\\
    \hline
    ZeroGen (Llama3) &  68.7 (4.0)& 73.8 (1.9) \\
    \hline
    ZeroGen (ChatGPT) &  60.2 (10.0)& 71.3 (1.5)\\
    \hline
    SuperGen (Zephyr)&  58.9 (5.0)& 65.8 (2.5)\\
     \hline
    SuperGen (Llama3)&  67.4 (2.4)& 71.4 (2.2)\\
     \hline    
    Refiner (Zephyr) &72.2 (5.3)&  76.0  (0.8)\\
    \hline
    Refiner (Llama3) &\textbf{72.4} (5.2)&  \textbf{76.9} (1.7)  \\
    \hline
  \end{tabular}
  \shrink

  \caption{Intent prediction accuracy is compared across unseen domains (averaged over 5 trials), with standard deviation reported in parentheses.  
  }
  \label{tab:other baseline}
\end{table}

\begin{table}[t]
  \centering
  \small
   \begin{tabular}{|l@{~}|c|c|c|c|c|}
    \hline
     & \multicolumn{2}{c}{SGD} & \multicolumn{2}{|c|}{Clinc150}  \\
     \hline

    & Dist-1 & Dist-2 & Dist-1 & Dist-2  \\
     \hline
    ZeroGen (Llama3) & 0.068& 0.163& 0.139 & 0.309 \\
    \hline
    ZeroGen (ChatGPT) &  0.080 & 0.213 & 0.147 & 0.348 \\
    \hline
    SuperGen (Llama3)& 0.058 & 0.131 & 0.120 & 0.259  \\
    \hline
    Refiner (Llama3) & \textbf{0.129}& \textbf{0.363} &\textbf{0.200} & \textbf{0.518}  \\
    \hline
    Real (human) Data  &  0.125& 0.378 & 0.167 & 0.436\\
    \hline
  \end{tabular}
  \shrink
  \caption{Lexical diversity results, averaged over 5 trials.}
  \label{tab:diversity}
\end{table}

\textbf{Effect of Refiner on unseen domains.} 
Table~\ref{tab:other baseline} shows that our approach outperforms all the baselines.
The table indicates that our method achieves better performance without the need for extensive sampling or using very large language models. 
The classifier trained with data generated from Llama3 performs better than that from Zephyr. After refinement, the refined data gives consistently better results, reducing the difference between Llama3 and Zephyr. As a comparison, a classifier trained on real (human) data is still substantially better, with an accuracy of 95.4\% in CLINC150. For SGD, we cannot make this comparison as all human data from SGD's unseen domains is used as test data, we do not have real human data for SGD to train a model for comparison.

\textbf{Lexical diversity and similarity.}
To validate our assumption that the refiner provides more diverse data, we report distinct-n as diversity metric. Table~\ref{tab:diversity} shows that the refiner leads to substantially more diverse data than other approaches. Additionally, the diversity of \textbf{SuperGen} is lower than of ZeroGen. 

\textbf{Qualitative analysis}
We manually compared the differences between the real data, the LLM-generated data, and the refined data. We found that refined data generally looks more like real data in terms of length and format. LLMs tend to make longer sentences because they often include the purpose or reason behind their intent. For example, for the intent of freezing an account, the real dataset includes the sentence ``i want my chase account blocked immeditately'', while an LLM output example is ``Can you put a hold on my account? I'm going on a trip and I don't want anyone to access my funds.''.

Additionally, LLMs sometimes provide unnecessary explanations for their generated utterance. For example, LLMs generate the utterance ``Can you please provide me with the phone number to text customer support? (User is asking for the specific contact information to send a text message to a customer support team)''. The refiner can remove these unwanted parts. However, the refiner can still occasionally generate incorrect utterances. For example, ``how do i report online'' is not clear enough to indicate the intent of reporting fraud. More details and examples are provided in Appendix \ref{sec:example}.

\section{Ablation Study}
\label{sec:abstudy}
\textbf{Effect of fine-tuning on performance.}
The results in Table~\ref{tab:other baseline} and \ref{tab:diversity} have shown that using refined language is better than directly using utterances generated by the LLMs. We investigate to what extent this improvement is due to the knowledge acquired from fine-tuning on seen domains. Table~\ref{tab:T5_FT_NFT} shows that the fine-tuned Refiner significantly improves overall accuracy over a non-fine-tuned Refiner, especially with a larger sample size. This confirms that the Refiner learns to improve utterances cross-domain.

\textbf{Factors to improve the training efficiency and performance.}
We additionally conduct an ablation study to explore various factors that could potentially affect training efficiency and performance:
1) We apply LoRA fine-tuning \cite{hu2021lora} with only around 5M trainable parameters to see if efficient fine-tuning can achieve similar outcomes. The results are in Table \ref{tab:LoRA}. We observe a slight drop in accuracy. 2) Our main results were based on using an input of 7 utterances. We experimented with different input/output settings to see if it affects performance. The results are in Table \ref{tab:inputoutput:unseen_PT}). We find that increasing the number of inputs slightly improves performance, while varying the number of outputs has minimal impact.

\begin{table}[t]
  \centering
  \small
  \begin{tabular}{|l@{~}|c|c|c|c|c|}
    \hline
    \multirow{2}{*}{Data source} & \multicolumn{2}{c|}{SGD} & \multicolumn{2}{c|}{Clinc150} \\
    \cline{2-5}
    & Zephyr & Llama3 & Zephyr & Llama3  \\
    \hline
    1$\times$   & & & & \\
    \hline
    Refiner (NFT) & 65.0  & 67.5  &73.3 & 73.8 \\
    \hline
    Refiner & \textbf{72.2}$^{**}$ & \textbf{72.4}$^{**}$ & \textbf{76.0}$^{**}$ & \textbf{76.9}$^{**}$ \\
    \hline
    2$\times$ & & & & \\
    \hline
    Refiner (NFT) &  62.9 &   67.8& 74.2 &  76.3\\
    \hline
    Refiner & \textbf{72.4}$^{**}$  &  \textbf{72.5}$^{**}$ & \textbf{77.2}$^{**}$&  \textbf{78.1}$^{**}$\\
    \hline

  \end{tabular}
  \shrink
  \caption{Intent prediction accuracy for unseen domains: with and without fine-tuning the Refiner (averaged over 5 trials). $**$ indicates the fine-tuned model significantly outperforms the non-fine-tuned (NFT) model at $\alpha = 0.05$ based on a one-tail paired t-test. 1$\times$ denotes the default evaluation sample size, and 2$\times$ denotes double.} 
  \label{tab:T5_FT_NFT}
\end{table}

\begin{table}[t]
  \small
\begin{tabular}{|l|c|c|c|c|c|}

    \hline
    \multirow{2}{*}{Data source} & \multicolumn{2}{c|}{SGD} & \multicolumn{2}{c|}{Clinc150} \\
        
    \cline{2-5}
    & Zephyr & Llama3 & Zephyr & Llama3  \\
    \hline
    Refiner      & \textbf{72.2} & \textbf{72.4} & \textbf{76.0} & \textbf{76.9} \\
    \hline
    Refiner (LoRA)        & 70.5  & 71.5 & 75.5 & 76.0 \\
    \hline

\end{tabular}

  \shrink
  \caption{Intent prediction accuracy for unseen domains (averaged over 5 trials): Full fine-tuning vs. LoRA parameter-efficient fine-tuning.
  }
  \label{tab:LoRA}
\end{table}

\begin{table}[!]
  \small
\begin{tabular}{|l|c|c|c|c|c|}

    \hline
    \multirow{2}{*}{Refiner(LoRA):} & \multicolumn{2}{c|}{SGD} & \multicolumn{2}{c|}{Clinc150} \\
        
    \cline{2-5}
    & Zephyr & Llama3 & Zephyr & Llama3  \\
    \hline
    1to1       & 69.7  & 70.5 & 74.0 & 75.1 \\
    \hline
    3to1       & 71.5  & 70.6 & 74.7 & 75.5 \\
    \hline
    5to1      & 71.0  & 72.1 & 74.1 & \textbf{76.8} \\
    \hline
    7to1    & 70.5  &71.5 & \textbf{75.5} & 76.0 \\
    \hline
    7to3     & 71.6 &  \textbf{73.4} & 74.9 & 76.7 \\
    \hline
    7to5    & \textbf{71.8}  & 71.8 & 74.1 & 76.0 \\
    \hline

\end{tabular}

  \shrink
  \caption{Intent prediction accuracy 
  for unseen domains (averaged over 5 trials) with varied input and output setting.
  }
  \label{tab:inputoutput:unseen_PT}
\end{table}

\section{Conclusions}
We found that (1) our proposed utterance Refiner can improve data quality on zero-resource domains without the need for larger LLMs or oversampling; (2) data from seen domains is still useful for generalizing to unseen domains; and (3) the Refiner enhances the lexical diversity of the LLM-generated data, comparable to human data. Our results indicate that a two-step approach of a generative LLM in zero-shot setting and a smaller sequence-to-sequence model can lead to high-quality data for intent detection. In future work, further work can be done to close the gap to the quality of human-labelled training data, which other models and other training strategies.

\section*{Acknowledgments}
This publication is part of the project LESSEN with project number NWA.1389.20.183 of the research program NWA ORC 2020/21 which is (partly) financed by the Dutch Research Council (NWO).

\section*{Limitations}

\textbf{Language.} Like most of the prior work, we only focused on English utterance generation. This is relevant because current LLMs are known to be more proficient in English than in other languages, and our results may not generalize to lower-resource languages.

\noindent
\textbf{Computational resources.} Not all settings that we originally wanted to experiment with were feasible with the computational resources that we had at our disposal (such as larger numbers of $n$ in the n-to-n settings).


\bibliography{anthology,custom}
\bibliographystyle{acl_natbib}

\clearpage

\appendix

\section{Domains for both datasets} 
\label{sec:domain}

The domains for each dataset are:
\begin{itemize}
\item CLINC150: Banking, credit cards, kitchen/dining, home, auto/commute, travel, utility, work, small talk, and meta.

\item SGD: Restaurants, media, events, music, movies, flights, ride sharing, rental cars, buses, hotels, services, homes, banks, calendar, weather, travel, alarm, payment, trains, and messaging.
\end{itemize}

\section{Distinct-n computation} 
\label{sec:eval}

We consider all utterances belonging to the same intent as a single text. For distinct-n, we calculate the ratio of unique n-grams to the total number of words in each document and then determine the average. Because distinct-n penalizes longer texts more, we follow ~\cite{Joko:2024:LAPS} and truncate the texts to the same length through sampling to ensure a fair comparison across generated datasets for each intent.

\section{Details of train-test split for the SGD dataset} 
\label{sec:split}

For the SGD dataset, there are 40 intents in the train-validation split and 34 in the test split, with 46 unique intents across both splits, of which 28 are shared. If we use the default splits, the unseen intents can only be selected from the 34 in the test split instead of all 46. To make our results more generalizable, we merge all splits together, allowing all intents to serve as possible evaluation. We randomly select 8 out of the 20 domains as unseen domains.  We train refiners on 9 ``seen\\ domains and monitor the validation loss on 3 ``seen'' domains.

We did not choose a 10/10 split because the SGD dataset has substantial fewer intents compared to the CLINC150 dataset (46 vs. 150), and our problem setting assumes we have sufficient seen-domain data.

Additionally, in SGD, the number of intents is not equally distributed across all domains. There are 4 domains with only 1 intent each, and 3 domains with 4 intents each. If we used a 10/10 domain split, it increase the chance of too few intents for the training split, undermining our assumptions. The number of examples for each intent ranges from 128 to 3,291, with an average of 1,244. Since each real human example has at least one unique generated pair from LLMs, and larger sample sizes would require more resources, which does not align with our low computing resource setting, we limit the training data to a maximum of 200 examples per seen-domain intent because some intents have fewer than 200 human examples, and we evaluate the refiners on unseen-domain intents.

\section{Details of the SuperGen baseline} 
\label{sec:baseline}

\paragraph{Applying SuperGen to our tasks.}
\citet{meng2022generating} use an LLM to generate 10$N$ examples for each unseen class. $N$ is the number of training examples. Subsequently, they calculated 
the geometric mean of the conditional generation probability for each example. The probability is conditioned on previously generated token(s) and the prompt. They selected $N$ examples with the highest score for training. Following their approach, we first generate $1,000 = 100 \times 10$ examples per intent for the Clinc150 dataset and $2,000 = 200 \times 10$ examples per intent for the SGD dataset. Then, we select the top 10\% confidence samples for evaluation. It's noteworthy that \citet{meng2022generating} also apply a regularization approach to train a classifier, but this aspect is not considered in our study.

\section{Some generated examples} 
\label{sec:example}
\paragraph{Examples of generated data}

We manually checked the generated data and there are some findings:

First, we found in the Clinc150, data generated by LLMs often produced longer utterances compared to real data. However, these utterances were shortened after refinement. Below are examples of the "freeze account" intent from the Clinc150 dataset\footnote{Some utterances are shown in their original format with lowercase at the beginning}:"

Real data
\begin{itemize}
    \item ``please freeze my account''
    \item ``i want my chase account blocked immediately''
    \item  ``i need you to block my chase account immediately''
\end{itemize}

Zephyr:
\begin{itemize}
    \item ``User: I would like to freeze my bank account temporarily.''
    \item ``User: Can you please put a hold on my bank account for a specific period of time?''
    \item ``User: I need to prevent any transactions on my account for a while.''
\end{itemize}

Llama3:
\begin{itemize}
    \item ``I want to freeze my account to prevent any unauthorized transactions.''
    \item ``Can you put a hold on my account? I'm going on a trip and I don't want anyone to access my funds.''
    \item ``I need to freeze my account temporarily until I can get a new debit card.''
\end{itemize}

Zephyr+Refiner
\begin{itemize}
    \item ``can you freeze my bank account until future notice''
    \item ``can you freeze my account''
    \item ``is there a way to freeze all my accounts''
\end{itemize}

Llama3+Refiner
\begin{itemize}
    \item ``i need you to stop opening my account''
    \item ``put my account on hold for now''
    \item ``"help, i am wanting my account to be frozen"''
\end{itemize}

Llama3+Refiner(LoRA)
\begin{itemize}
    \item ``freeze my account from now on''
    \item ``help, i am trying to freeze my card''
    \item ``can you do frozen transactions please''
\end{itemize}

Secondly, we observed that in the refined data, new words were used to express the intent. Below are examples of the refined output from Llama3:
\begin{itemize}
    \item ``get your approval to immediately halt all activity on my account (intent: freeze account): The word `halt' is not used in the Llama3 output''
    \item ``can you tell me the steps for handing in the money to my mom's account (intent: transfer): The phrase `hand in' is not used in the Llama3 output''
\end{itemize}

Thirdly, we observed that the Refiner filtered out some noise words. For instance, in Zephyr's outputs, there were instances where "User:" was frequently used as a prefix, and occasional extra explanations were provided (see examples below). These occurrences were reduced after refinement.
\begin{itemize}
    \item ``Can you please provide me with the phone number to text customer support? (User is asking for the specific contact information to send a text message to a customer support team)'' (intent: text)
    \item ``User: I want to set a timer for 60 minutes, but I don't want it to beep when it's done. Is that possible? (Note: This utterance also indicates a preference or request for customization.)'' (intent: timer)

\end{itemize}
Fourthly, we found that the Refiner with LoRA fine-tuning generates incorrect utterances more frequently than the fully fine-tuned Refiner. Below are examples selected from both datasets, with the correct intent shown in parentheses:

\begin{itemize}
    \item ``help with my credit cards'' (intent: freeze account)
    \item ``how do i report online'' (intent: report fraud)
    \item ``please help'' (intent: account blocked)
    \item ``I want to get a cab to take me somewhere from the city of San Francisco.'' (intent: find bus)
\end{itemize}

\end{document}